# Coupled Physics-Gated Adaptation: Spatially Decoding Volumetric Photochemical Conversion in Complex 3D-Printed Objects

## Submitted for consideration by Additive Manufacturing

**Authors:** Maryam Eftekharifar[1], Churun Zhang[2], Jialiang Wei[2], Xudong Cao[2], Hossein Heidari*[2]

[2]*Institute for Materials Discovery, University College London, London E20 2AE, UK*

[1]*School of Biosciences, University of Birmingham, Edgbaston B15 2TT, UK*

**Corresponding author (h.heidari@ucl.ac.uk)*

A framework is presented that pioneers the prediction of photochemical conversion in complex three-dimensionally printed objects, introducing a challenging new computer vision task of predicting dense, non-visual volumetric physical properties from 3D visual data. This approach leverages the largest-ever optically printed 3D specimen dataset, comprising a large family of parametrically designed complex minimal surface structures that have undergone terminal chemical characterisation. Conventional vision models are ill-equipped for this task, as they lack an inductive bias for the coupled, non-linear interactions of optical physics (diffraction, absorption) and material physics (diffusion, convection) that govern the final chemical state. To address this, we propose Coupled Physics-Gated Adaptation (C-PGA), a novel multimodal fusion architecture. Unlike standard unimodal vision and multimodal concatenation, C-PGA explicitly models physical coupling by using sparse geometrical and process parameters (e.g., surface transport, print layer height) as a Query to dynamically gate and adapt the dense visual features via feature-wise linear modulation (FiLM). This mechanism spatially modulates dual 3D visual streams extracted by parallel 3D-CNNs processing raw projection stacks and their diffusion-diffraction corrected counterparts allowing the model to recalibrate its visual perception based on the physical context. This approach offers a breakthrough in virtual chemical characterisation, eliminating the need for traditional post-print measurements and enabling precise control over the chemical conversion state. Experimental results demonstrate state-of-the-art predictive fidelity, achieving a Pearson correlation of 0.8931 against ground truth and accounting for over 77% of the variance with a low mean absolute error of 0.0200. Ablation studies confirm the necessity of our architectural design: a numerical-only model fails completely, while C-PGA dramatically outperforms an image-only baseline, achieving a 48% relative increase in explained variance. Crucially, we move beyond aggregate metrics to validate the model as a scientific instrument. Extensive interpretability tests, including perturbation-based occlusion sensitivity analysis, provide strong visual evidence that the model's predictions are based on scientifically salient features at the material-void interface, precisely where diffraction and diffusion effects are most pronounced. This rigorous validation framework establishes a new benchmark for developing trustworthy, physics-aware vision models for advanced manufacturing.

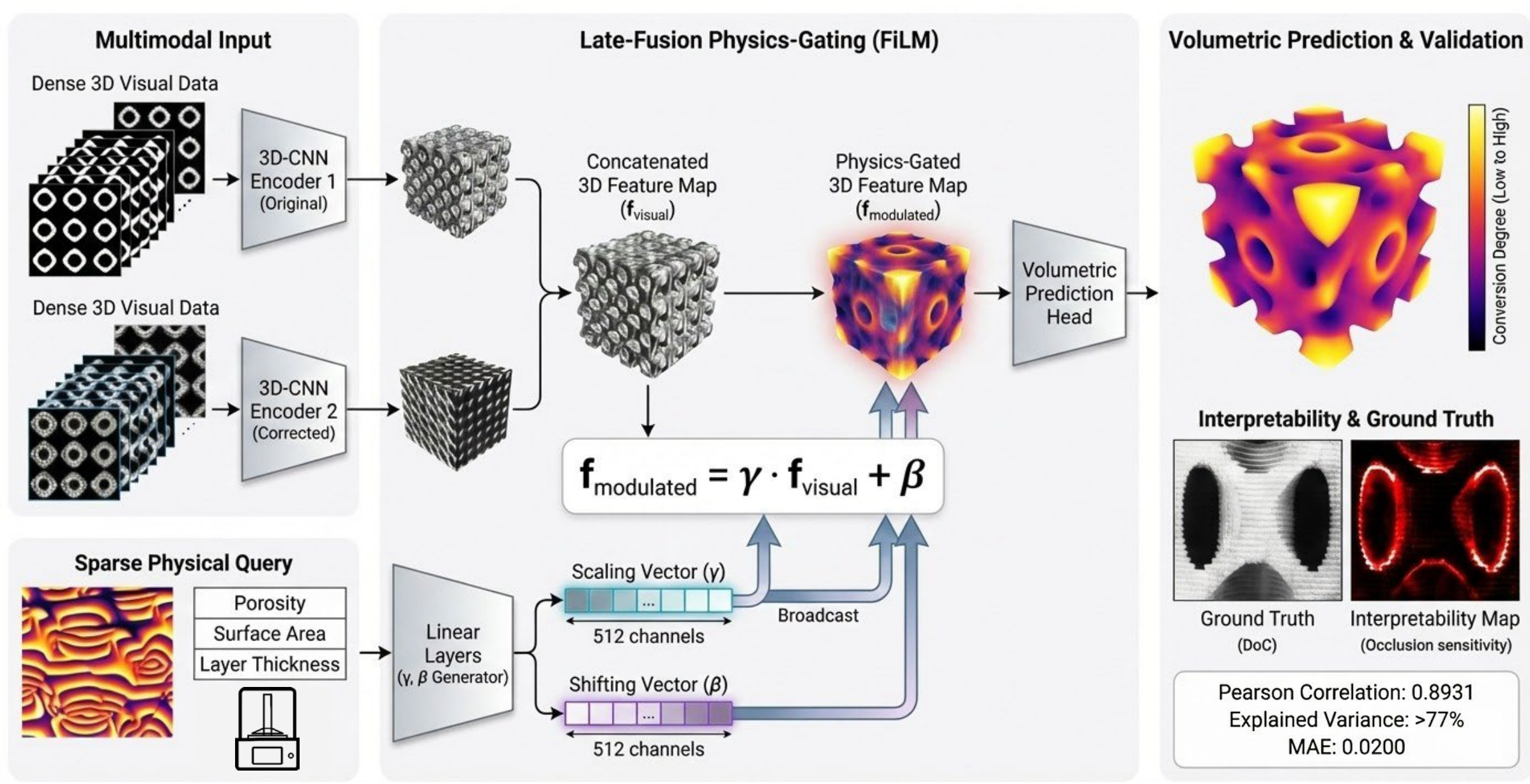

## 1. Introduction

Light-assisted additive manufacturing also referred to as stereolithography (SLA) is commonly represented by 2D processes such as vat photopolymerisation and its subvariants such as continuous liquid interface production (CLIP[1]), mobile liquid interface production (HARP[2]) and xolography[3] where the processing unit is two-dimensional layers, there are also point-to-point (0D) processes such as two-photon lithography (TPL[4]) and more recently volumetric (3D) processes such as computed axial lithography (CAL[5]) and holographic lithography (HoloTile[6]) where volumes of liquid precursor are processed concurrently. With the introduction of multifunctional high-performance photopolymers, optics-enabled superior print resolutions, improved repeatability, speed and throughput, and liquid phase manufacturing flexibility, all variants of SLA processes have received major traction.

Process modelling has been a key contributor to the accelerated growth of such light-based additive manufacturing systems. An emerging approach in this regard is machine learning. Machine learning is already transforming additive manufacturing, enabling unprecedented levels of process control, quality assurance, and materials design. Recent work in the intersection of machine learning and light-based additive manufacturing has moved rapidly from passive monitoring to prediction, mask design, and multi-objective optimisation. There is a clear shift from data-driven prediction towards sophisticated, physics-aware, and generative models deeply integrated into the entire additive manufacturing workflow.

Seminal work by Scime et al. pioneered the use of convolutional neural networks (CNNs) for *in situ* defect detection, demonstrating an innovative architecture capable of identifying anomalies like powder bed spreading errors from raw image data in real-time[7]. Another early demonstration in light-based additive manufacturing by Lee et al. framed *in situ* TPL videos as a three-state quality classification task (uncured, cured and damaged), showing that compact CNNs trained on curated process videos can automate exposure tuning and flag damage in real time, seeding open datasets for the field[8]. Building on this, Hu et al. used dual visual streams explicitly incorporating spatiotemporal networks (3D-CNN and CNN-LSTM) to couple 'during-print' video with 'post-process' microscopy, improving parameter search speed while capturing polymerisation dynamics that 2D models miss as an architectural step forward that treats printing as a computer vision problem[9]. Challenges such as scattering compensation were addressed by You et al. casting inverse mask designs as image-to-image learning with a U-Net-style encoder–decoder, generating grayscale masks that counteract turbid-resin scattering and measurably boost print fidelity[10]. Pingali et al. trained classification networks on reaction–diffusion simulation data to create fast surrogate 'printability' predictors for TPL, a pragmatic physics-generated data route that avoids costly experiments while retaining mechanistic signal[11].

While print fidelity as defined by geometrical accuracy and error, has been the core focus of most studies, the primary challenge facing SLA remains to be predictability and control over transport- and kinetics-driven volumetric chemical characteristics such as photochemical conversion. This is highly challenging when objects are emerging, property maps dynamically changing and boundaries shifting throughout the process. While all mentioned SLA processes rely on some form of post-processing and post-print flood exposure or baking to ensure terminal properties are achieved, a powerful model to predict an arbitrary object's bulk properties using processing parameters, projection image sets and material composition is highly desirable. In the case of all the abovementioned projection-based SLA methods, while processing parameters such as dark photopolymerisation time and fluid motion are involved, the spatial exposure intensity projection sets, exposure intensity-time profile, and processing layer or virtual layer thicknesses are the most significant determinants of final properties. Notably, the chemical kinetics are influenced by the dynamically emerging geometrical features of the printed object that govern proximity effects, light propagation, chemical transport and heat transfer in the spatiotemporal domain. Attempts to model and predict the outcome of light-assisted additive processes have been highly restricted and limited to simplified two-dimensional models, convolution models and oversimplified reaction kinetic models that fail to capture the full interplay of factors involved in the three-dimensional process[12–14]. A common indicator of the overall outcome of such processes is the degree of chemical conversion (DoC), which remains highly unpredictable for the resulting 3D object. The DoC is an indicator of the percentage of monomer molecules transformed into the crosslinked polymer state, representing the degree to which the reaction has been completed across the object. This is a key factor in determining the suitability of the printed part for consumer use, medical implantation and long-term stability and chemical leaching, or the need for further post-processing steps[15].

To map a reasonably large geometrical space, we would require a systematic sweep of the domains of architectural parameters that influence light propagation, chemical transport and thermal transport. These parameters include 3D feature proximity, surface area availability, porosity and feature size. The approach of choice for convenience and automation was to map the commonly incorporated framework of cellular porous lattices. Such geometries have been widely used in multiple engineering domains and identify with random, mathematical or generative repeating patterns of lattices, and in the case of the study, allow for scalable sample generation and parameter tuning. For analogue reasons, they are used widely for their favourable tuneable mechanical performance and surface availability in tandem, making them perfect candidates for biomedical and chemical engineering applications[16].

Common examples of 3D cellular lattices include strut-based (formed by bulk rods or struts connected in nodes[17]) and triply periodic minimal surface-based (TPMS, aligned by a series of non-self-intersecting periodic curve surfaces[18]). The latter is classified into skeletal and sheet lattices, two groups using the same TPMS single sheet but implementing different construct methodologies. The mechanical, transport and thermal properties, relative density, energy absorption and other performance metrics vary among these various scaffolds. Each TPMS lattice is generated with a unique trigonometric function with edge-free, interconnected final surfaces. This tunability allows them to be easily designed, optimised and fabricated with various distinctive features including unit cell size, pore size, thickness, porosity, and surface area to volume ratio. These kinds of surfaces occupy the smallest area within a given three-dimensional space, termed a 'catenoid'. Catenoids can be described by parametric equations

$$x = c \cosh\left(\frac{v}{c}\right) \cos u \tag{1}$$

$$y = c \cosh\left(\frac{v}{c}\right) \sin u \tag{2}$$

$$z = v \tag{3}$$

Yielding the following Fourier series of theoretical trigonometrical equations

$$\varphi(r) = \sum A_k \,{}_{k=1}^{K} \cos[2\pi(h_k \cdot r)/\lambda_k + p_k] = C \tag{4}$$

Where $r$ is the location vector or can be shown as $(x, y, z)$, $A_k$ stands for the amplitude factor, $h_k$ represents the lattice vector in reciprocal space, $\lambda_k$ describes the wavelength of periodic repeating pattern, $p_k$ is the phase shift and C is a constant which controls the thickness of generated TPMS structures. Shwartz and his group adopted and improved the TPMS lattice structure for the first time in 1890 and pushed this forward to fit more different structures including Diamond, Neovius, and Fischer-Koch lattices. After a century, Schoen introduced a new frame called the gyroid, one of the most investigated topics in the field at this time[16]. These different spaceframes can also be generated using minimal surface functions. The TPMS family keeps expanding with approximately 100 lattices to date. Cellular structures in the present study were generated using the level-set approximation equations (refer to SI, Eqn. S1-S6).

The C value, also referred to as the offset parameter, is definitive in determining the optimal geometry parameters, including porosity, relative density, and surface area. These are parameters with a critical influence on chemical conversion dynamics through their influence on spatiotemporal reaction propagation and molecular transport. In skeletal-based structures, the significant shift point for C is observed at the zero point, whereas in sheet-based cases, the defined range is of greater consequence. The necessity for a range of C values differs depending on the geometry in question. Here, the wide range of possible C values was narrowed in the implementation process to the printable range which is reported. Furthermore, the construction of sheet-based structures by restricting two opposing offset surfaces, results in the formation of two distinct thicknesses in the corner regions[19]. This may potentially lead to nonuniformities across the structure, further causing deficiencies of both local and global self-intersections. Consequently, an enhanced methodology was used to construct the lattices, whereby the minimal surface is offset in the direction of the normal vector, forming isolayers which are later closed with a boundary. This was implemented in the present study (refer to SI section 2, Eqn. S7-S17).

## 2. Results and discussion

A multimodal deep learning architecture is developed to predict DoC in vat-photopolymerised objects. The models are evaluated with a three-part performance–uncertainty–mechanism framework tailored for scientific deployment. We first establish predictive fidelity on held-out data, comparing the learned distribution of DoC to experimental measurements and analysing error structure and residual behaviour to assess generalisation robustness. Next, we move from aggregate metrics to reliability characterisation, constructing spatial maps of model uncertainty across the design space to delineate where predictions are most dependable and where additional data or tighter process tuning may be required. Finally, we interrogate the model's decision process with complementary attribution analyses using gradient-based saliency for rapid screening and perturbation-based occlusion for counterfactual testing to assess whether the learned representations align with physically meaningful features rather than spurious cues. Together, these components situate the model as a candidate scientific instrument; the following sections present quantitative performance, spatially resolved uncertainty, and interpretability results in turn. While the integration of machine learning with physics-aware modelling holds immense promise for accelerating materials discovery, the black-box reputation of deep networks remains a barrier to adoption. Hence, we target high predictive accuracy in conjunction with scientific plausibility and mechanistic consistency.

### 2.1. Experimental dataset of printed 3D objects

In this study six representative cellular architectures were selected, namely Schwarz Primitive, Diamond, Fischer-Koch, Gyroid, F-RD and Neovius lattices. A total of 648 configurations were generated, meshed and sliced (Fig 1a). As previously stated, each extended surface is defined using level-set approximation trigonometric functions. Two variables for the topology parameter were selected: unit cell number and the offset C value. Each of these variables are assigned four and three distinct values respectively (Fig 1f), allowing for a full spectrum indirect modulation of porosity, specific surface area, proximity and feature size. In contrast with the methodology usually employed in previous experiments, the present study adopted a direct approach to introducing the irradiance parameter through the arrangement of structures on the print platform to allow for such large numbers of specimens to be printed. A linear decrease in irradiance as a function of distance from the center of the vat membrane was utilised as shown in Fig 1e.

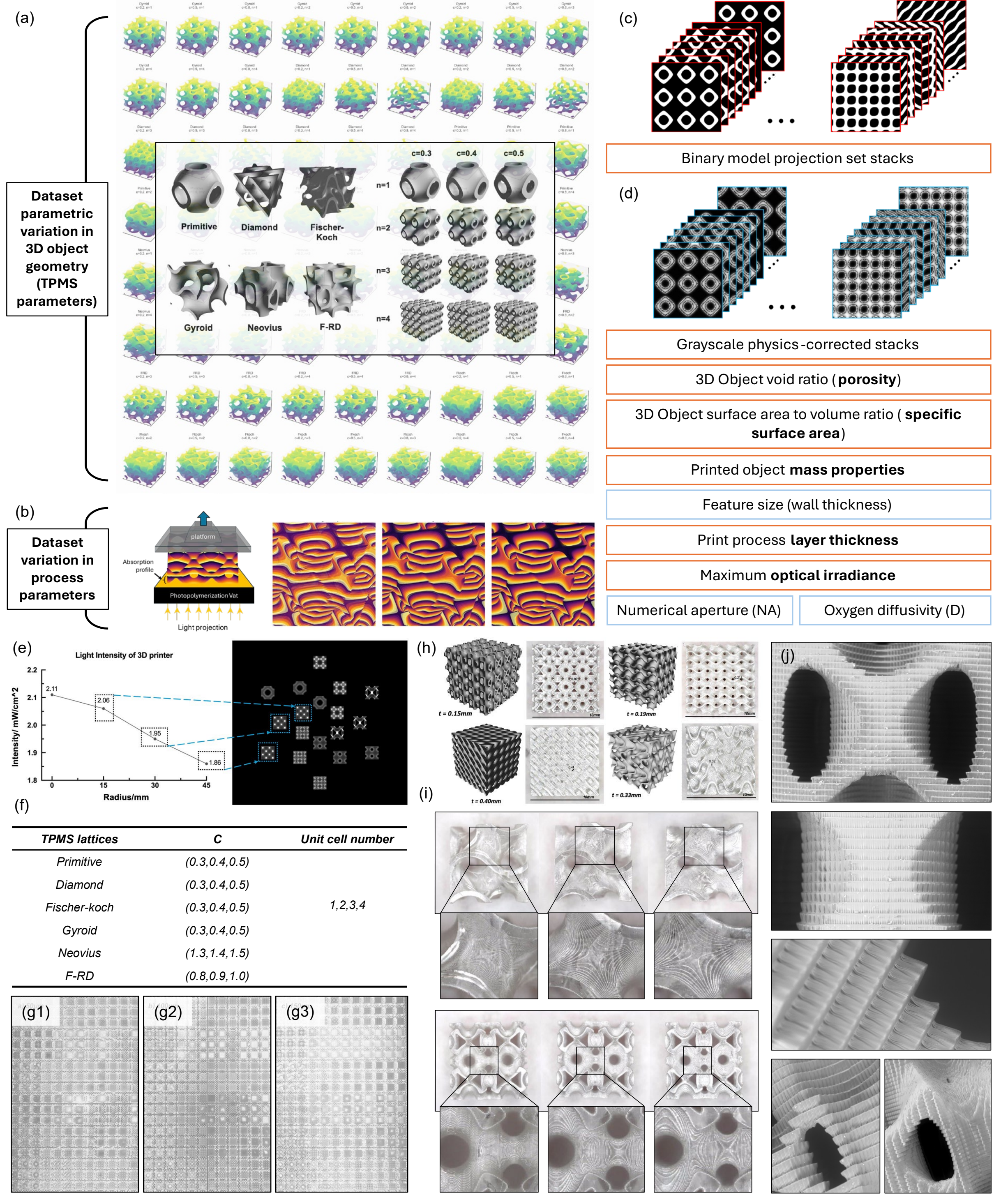

| *TPMS lattices* | *C* | *Unit cell number* |
|---|---|---|
| *Primitive* | *(0.3,0.4,0.5)* | |
| *Diamond* | *(0.3,0.4,0.5)* | |
| *Fischer-koch* | *(0.3,0.4,0.5)* | *1,2,3,4* |
| *Gyroid* | *(0.3,0.4,0.5)* | |
| *Neovius* | *(1.3,1.4,1.5)* | |
| *F-RD* | *(0.8,0.9,1.0)* | |

**Figure 1** *C-PGA training dataset, and the optical and geometrical parameters utilised to construct the dataset: (a) Architectural and geometry-related features and a three-dimensional representation of the lattices; (b) Physical and process-related features and a visualisation of optical irradiance across layers of the construct assuming basic beer-lambert law. A schematic of the optical configuration with a three-dimensional dosage heatmap; (c) Projection stacks used as the key image modality; (d) transformed image stacks incorporating a convolution with diffusion and diffraction kernels; (e) The irradiance variation with distance for the radially arranged TPMS constructs at three different radial positions per geometry; (f) Geometrical parameter set for generated TPMS structures; (g) Display grid of all printed specimen (x-axis representing 3 thicknesses × 4 unit cell numbers × 3 print configurations; Y-axis: 3 irradiance values × 6 TPMS geometries); (h) Examples of TPMS architectures displaying original design and actual print cross-section; (i) Close-up micrographs of complex staircasing artifacts in three different print layer height configurations for two of the TPMS geometries; and (j) Scanning electron micrographs of the layers and the structural intricacies caused by the naturally pixelated projections.*

Hence, the print parameters are defined as the combination of layer thickness and irradiance. Each print operation containing 18 single TPMS structures was sliced and repeated with three different layer thicknesses: 50, 100 and 150 µm (Fig S2). A comprehensive set of print parameters implemented in these experiments including the core parameters above, the initial exposure time, turn off delay, rising height, platform pump velocity, and number of base layers are reported in the supplementary information (refer to SI, Table S1). A simulated calculation of the surface area and porosity of the lattice structures was carried out. Surface area encompasses all surfaces of the porous structure that are in contact with the surrounding space. Porosity or volume fraction represents the percentage of the generated area occupied by empty space. The observed cracks associated with the solvent extraction process were induced during the drying phase and hence were not involved in SA/V calculations as they were not present in surface-mediated processes during print or post-processing (Fig S1). It is evident that both indirect parameters exert a significant influence on surface-mediated processes, transport, conduction and convection. Details of the algorithmic approach for calculation of each parameter are provided (refer to SI section 3).

### 2.2. High-fidelity prediction of photochemical conversion degree through multimodal learning

The proposed C-PGA framework demonstrates a remarkably strong capacity to predict the volumetric degree of conversion, accurately capturing the target variable's distribution across the entire test set. As shown in the regression analysis (Fig 2), the predictions of our final late-fusion FiLM C-PGA model span the experimental range from approximately 0.67 to 0.95, closely mirroring ground truth values. This high fidelity is quantitatively substantiated by a strong Pearson correlation coefficient (PCC) of 0.8931 and a coefficient of determination ($R^2$) of 0.7679, indicating that the model accounts for nearly 77% of the variance in the complex physicochemical data.

An analysis of the error profile reveals exceptional precision and stability. The late-fusion FiLM model achieves a low mean absolute error (MAE) of 0.0200 and a root mean square error (RMSE) of 0.0279. Crucially, the residual plot (Fig 2) demonstrates that these errors are tightly and randomly distributed around the zero-line, lacking the systematic bias observed in simpler models. The training dynamics (inset in histogram) illustrate efficient convergence, with validation loss tracking training loss closely, confirming robust generalisation without overfitting. To isolate the drivers of this performance, we performed a rigorous ablation study across six architectural variants (Fig 2, rows 1-6). The necessity of a multimodal approach is irrefutably confirmed by the failure of unimodal baselines. The Numerical-only model exhibits a catastrophic failure, achieving an $R^2$ of approximately 0 and severe systematic bias (residuals spanning -0.5 to +0.4), proving that process parameters alone cannot capture the spatially dependent physics. The image-only model performed significantly better (PCC = 0.7407), confirming that geometry is a dominant factor, yet remained imprecise with an MAE of 0.0277 and low $R^2$ of 0.5187.

The integration of visual and physical modalities yielded a substantial performance leap. The initial concatenated multimodal baseline (row 3) jumped to an $R^2$ of 0.7572, representing a 46% relative increase in explained variance over the image-only model. However, our C-PGA architectures pushed this boundary further. While the hierarchical FiLM model (row 5) achieved the highest raw metrics ($R^2$ = 0.7806), our analysis prioritised trustworthiness alongside accuracy. As detailed in the interpretability section, the hierarchical model's decision-making proved opaque. The late-fusion FiLM C-PGA (row 6) provided a comparable high-fidelity prediction ($R^2$ = 0.7679) and error distribution. Both FiLM-based models have a regression fit (red line) visibly tighter to the ideal diagonal than Attention-based C-PGA (row 4, $R^2$ = 0.7539), particularly at the critical lower-conversion extremes. This confirms that the explicit spatial gating provided by the FiLM architectures could be the optimal strategy for coupling sparse physical parameters with dense 3D visual data.

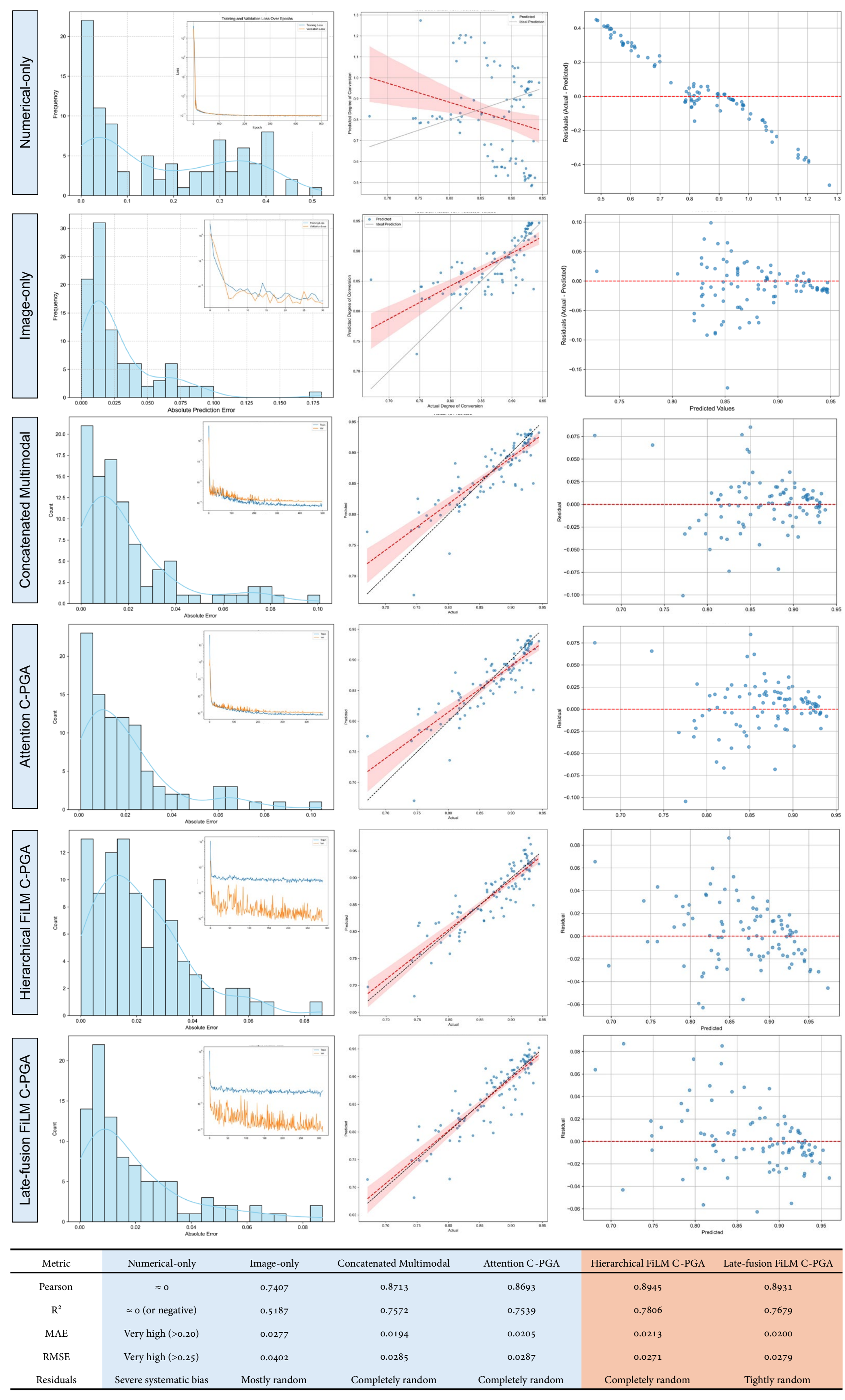

| Metric | Numerical-only | Image-only | Concatenated Multimodal | Attention C-PGA | Hierarchical FiLM C-PGA | Late-fusion FiLM C-PGA |
|---|---|---|---|---|---|---|
| Pearson | ≈ 0 | 0.7407 | 0.8713 | 0.8693 | 0.8945 | 0.8931 |
| $R^2$ | ≈ 0 (or negative) | 0.5187 | 0.7572 | 0.7539 | 0.7806 | 0.7679 |
| MAE | Very high (>0.20) | 0.0277 | 0.0194 | 0.0205 | 0.0213 | 0.0200 |
| RMSE | Very high (>0.25) | 0.0402 | 0.0285 | 0.0287 | 0.0271 | 0.0279 |
| Residuals | Severe systematic bias | Mostly random | Completely random | Completely random | Completely random | Tightly random |

**Figure 2** *Design and performance analysis of all unimodal and multimodal models: basic performance analysis on the test set displaying actual vs predicted values of photochemical conversion; distribution of absolute errors; plot of residuals; and plot of training and validation loss over epochs. The summary table highlights the effectiveness of various fusion strategies.*

### 2.3. Spatial mapping of model uncertainty via latency analysis

While aggregate performance metrics establish the model's overall accuracy, the latency maps (Fig 3) allow for a granular, quantitative analysis of predictive uncertainty across the input parameter space. This analysis transforms the model from a passive predictor into an active instrument for probing the underlying physics of vat photopolymerisation, highlighting regions of heightened sensitivity and potential manufacturing instability. The maps reveal that prediction error is not uniformly distributed but is concentrated in specific, physically meaningful regimes. Comparing the architectural variants reveals that while the concatenated and attention models (rows 1 & 2) exhibit sporadic and randomly-spread high-error clusters (yellow/orange points), the FiLM-based C-PGA models (rows 3 & 4) display localised regions of increased error while maintaining a consistently low error profile (predominantly dark purple) across the rest of the design space.

The middle and right column plots reveal a distinct cluster of higher error for structures with a void ratio below 0.7 and SA/V ratio above 1500. This corresponds to lattice configurations with extremely thick struts and smaller pores. In this regime, physical structures become highly susceptible to stochastic manufacturing variations. Owing to the larger solid regions acting as localised inhibitor-depleted zones of polymerisation, overcuring beyond the desired contours can occur as result of the photopolymerisation front with low conversion growing rapidly. Hence, minor fluctuations in irradiance or localised inhibitor concentration have a disproportionately large impact on the final geometry. Notably, while FiLM-based models (especially late-fusion) capture this effect in a more localised manner, this uncertainty persists across all model architectures (yellow/orange cluster in top-left region, middle-column plots), validating that this is a fundamental physical limit of the manufacturing process itself a stochastic noise floor rather than a deficiency in the C-PGA architecture.

The identification of such a critical region of uncertainty for samples with a combination of high SA/V ratio and low void ratio (thicker, bulkier material sections with still relatively high exposed surfaces), demonstrates signs of model interpretation of the heightened sensitivity to the effects outlined above, at sections with increased transport. Here, the distinction between architectures becomes evident. The concatenated model (row 1) struggles significantly, showing several high error outliers. This likely stems from its inability to model the complex, non-linear curing dynamics—such as inhibitor accumulation and quencher diffusion that dominate in bulkier features. In contrast, the late-fusion FiLM model (row 4) significantly dampens these errors. Its physics-gating mechanism effectively adapts the visual features to account for these depth-dependent bulk effects, yielding a more reliable prediction even in kinetically constrained regimes.

The left-column plots demonstrate the model's stability across processing parameters. The distribution of low-error (purple) points indicates that all multimodal architectures effectively decouple the influence of layer height from geometric complexity. This granular analysis confirms that our physics-gated architecture does not merely fit the data but successfully captures the governing constraints of the photopolymerisation process, providing invaluable feedback for process optimisation. To validate that the model's predictive capabilities are rooted in scientifically relevant features, we employed a dual-pronged interpretability strategy. In the following sections, we use a gradient-based method (saliency maps) for a computationally efficient, first-pass analysis, complemented by a more robust, perturbation-based method (occlusion sensitivity) for rigorous validation. Expanded results from multiple, diverse samples can confirm whether or not the C-PGA model has learned a generalisable and physically grounded feature representation.

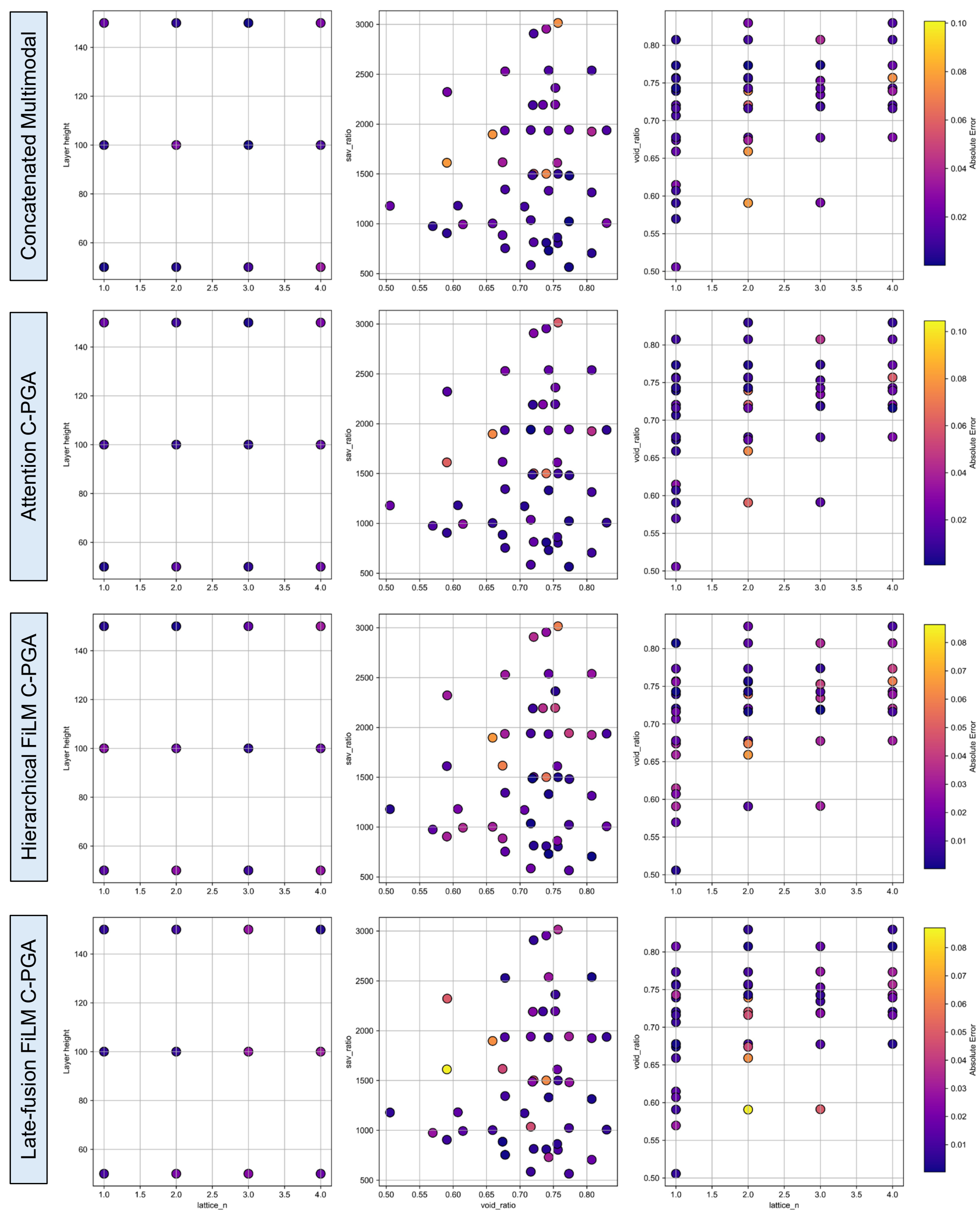

**Figure 3** *Spatial analysis of predictive uncertainty (latency maps) across the architectural ablation study. A comparative visualisation of the absolute prediction error projected onto 2D slices of the input parameter space for four distinct model architectures: Concatenated multimodal baseline, Attention C-PGA, hierarchical FiLM C-PGA, and late-fusion FiLM C-PGA. The columns represent key interactions between manufacturing and geometric parameters: (left) layer height vs. lattice N; (middle) SA/V ratio vs. void ratio; and (right) void ratio vs. lattice N. The colour scale represents the magnitude of the absolute error, where dark purple indicates high precision (<0.02) and yellow indicates high uncertainty (>0.10).*

## 2.4. Spatial analysis to validate salient learning

The raw saliency maps (Fig 4, left columns), derived from the magnitude of back-propagated gradients, exhibit high-frequency spatial components, a known characteristic of this technique which can amplify sensitivity to small textural variations in the input. Despite this, the overlay plots (right columns) provide a clear, quantitative insight. In all samples, regions of high feature attribution, with a normalised saliency value > 0.8 (yellow), consistently co-locate with the solid-phase material of the lattice struts and nodes. Conversely, the void phase corresponds to regions of low attribution, with saliency values < 0.2 (dark purple). This quantitatively demonstrates that the model's gradient is maximally sensitive to the presence of the polymerised material. This impressive performance is valid even for the finest structures and smallest pore sizes as can be seen in some of the finer lattices in Fig 4. While this confirms the model is attending to the correct material phase, further insight can be gained from contour analysis below and a more direct, perturbation-based validation in the following section.

The most critical validation of the proposed framework comes from interrogating its internal decision-making process across the architectural ablation study. Figure 4 presents a comparative multi-panel saliency analysis for sixteen random lattice geometries. This visualisation reveals a divergence in feature attribution behaviour, directly motivating our selection of the late-fusion FiLM C-PGA as the optimal architecture. The hierarchical FiLM model, despite its high quantitative metrics, produces diffuse and jittery saliency maps. This scattering suggests a reasoning process that is highly distributed across the network depth, making it opaque and difficult to interpret as a scientific instrument. In sharp contrast, late-fusion FiLM C-PGA produces remarkably clean, high-frequency maps with the highest signal-to-noise ratio. Its gradients are sharply localised, providing the clearest window into the physics captured by the model. Coupled with other elements of our analysis where the FiLM-based C-PGA models excel, this essential study confirms the suitability of the proposed model.

Focusing on the late-fusion results, a consistent physical logic emerges. In all samples, regions of high attribution (normalised saliency > 0.8) strictly coincide with the solid phase material, while void phases are correctly suppressed (< 0.2). More importantly, the model is not simply detecting mass; it is highly sensitive to the material-void interfaces, precisely where optical diffraction and chemical diffusion gradients are most pronounced. This physics-aware attention adapts dynamically to the topology. In the case of thick strut lattices, present in structures with large, open cells, saliency is not uniform across the strut but concentrates at the inner and outer boundaries. This indicates the model has learned that the cure state at the material-resin interface where inhibitor diffusion is active is a critical predictor of bulk conversion. This is while for fine feature lattices present in dense lattices with thin struts, the saliency map lights up the entire strut network. In these regimes, internal transport lengths are shorter and also diffraction zones from adjacent features overlap significantly, making the entire solid volume a region of high physical interaction and, thus, high importance.

For conventional node-based lattices, the model places maximal importance on the junctions and curved sections. These are areas where anisotropic transport and diffusion effects create highest variability in photopolymerisation, and the model has correctly identified them as information-rich regions. Conversely, the deep interior of very thick struts with stable conversion, and empty voids are consistently assigned low saliency. This confirms that the late-fusion FiLM architecture has moved beyond simple pattern matching and has learned to focus on the geometric complexities that drive the underlying physics of the manufacturing process.

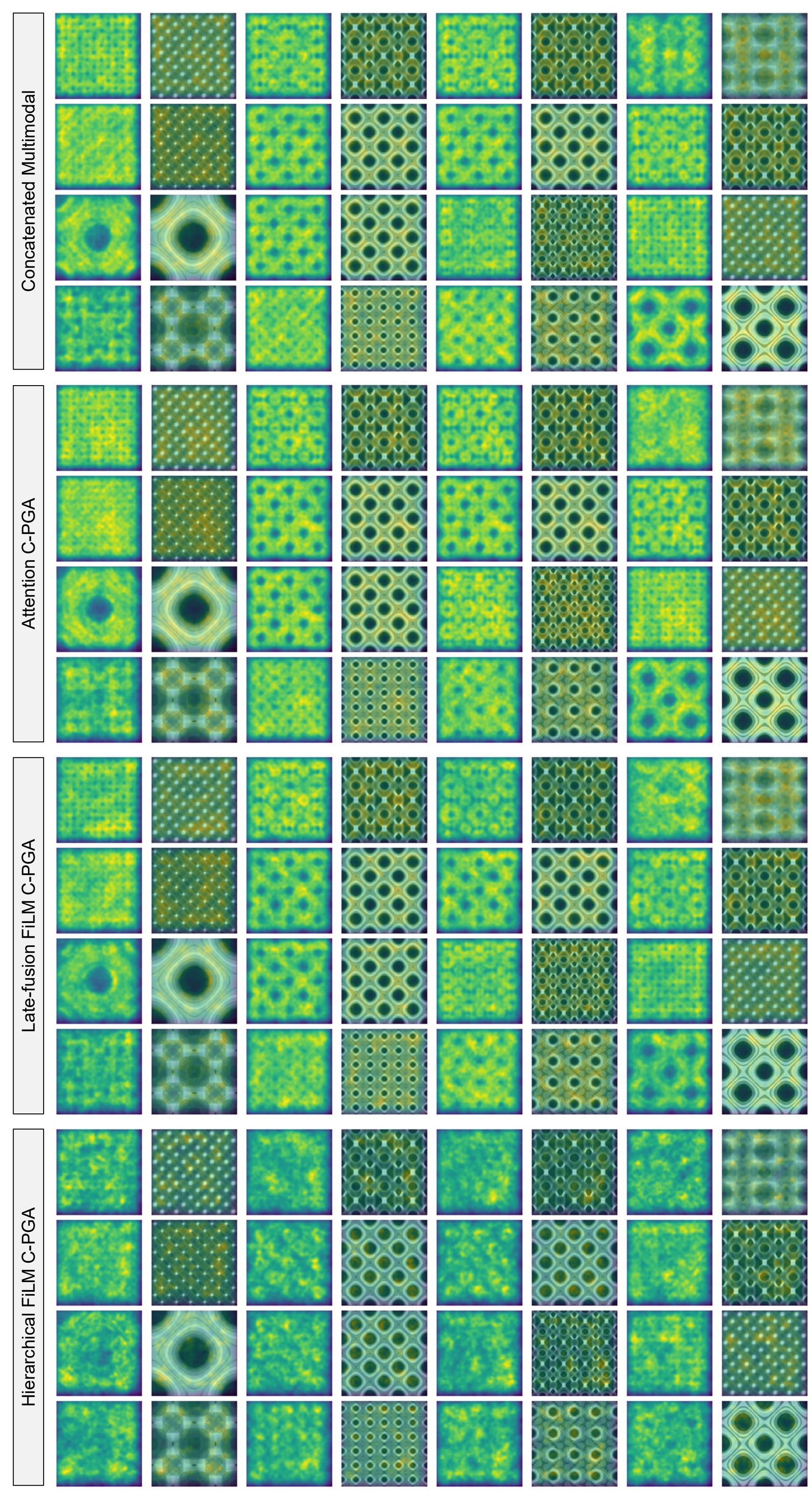

**Figure 4** *Spatial analysis of model uncertainty: Saliency map analysis for 16 random test specimens including for each specimen, the map and image slice overlayed map. All four multimodal models are compared side-by-side.*

## 2.5. Validation of scientifically salient learning through occlusion sensitivity analysis

The occlusion sensitivity analysis serves as a key ground-truth validation of the model's learned causal reasoning. Figure 5 presents a comparative counterfactual analysis across the two FiLM-based C-PGA architectural variants. By systematically perturbing (occluding) local regions of the input and measuring the absolute change in predicted conversion, we generated high-resolution sensitivity maps for 16 randomly selected samples representing a diverse range of lattice topologies. The visualisation reveals a striking difference in feature localisation capabilities. While gradient-based saliency maps often yield diffuse, noisy visualisations due to the vanishing gradient problem in deep 3D architectures, the perturbation-based occlusion analysis presented here provides a discrete, high-fidelity map of the model's decision-making topology. The following analysis suggests that the FiLM-based C-PGA models have learned a physics-based dependency and contextual inference, where the model derives the state of a solid feature not by looking at the feature itself, but by analysing the surrounding environment (the void) that dictates its chemical fate.

A first observation from the hierarchical FiLM occlusion test is the counter-intuitive inversion of saliency observed in finer lattice structures (tests 2 and 13). Unlike standard computer vision tasks where the object of interest typically triggers the highest activation, here the model explicitly treats the void space as the governing variable for prediction, and there is a clear negative correlation between the physical solid and the model's visual attention. This is a strong signature that the model is emphasising oxygen inhibition effects where the conversion of a voxel is rate-limited by the diffusion of oxygen (inhibitor) from the surrounding voxels. If the void is occluded, the model's calculation of the diffusive flux entering the strut is influenced, causing a massive drop in prediction accuracy. In comparison with the late-fusion model, despite high aggregate metrics, the hierarchical model exhibits blocky and diffuse sensitivity maps.

The late-fusion occlusion sensitivity results similarly reveal an intriguing C-PGA model behaviour. In thick structures (late-fusion tests 7 and 9), the model seems to learn to pay attention (higher heatmap temperatures) to the solid core, while for thin structures (late-fusion tests 13 and 15), the model looks at the interfaces and void regions. As further explained below, the results demonstrate that the proposed C-PGA models transcend simple segmentation. They do not merely look for where the material is located but instead identify where the reaction kinetics are determined.

For bulk-dominated regimes the model focuses on solid mass in structures with low specific surface area and large nodal domains (Neovius-like or primitive geometries, test 7 and test 9) and the heatmaps show a strong positive correlation between the solid phase and predictive importance. The brightest hotspots in the heatmap overlap almost perfectly with the thickest white regions of the projection. In these bulky regions, oxygen diffusion from the surrounding resin is negligible due to the large diffusion distance to the center of the node. Consequently, the conversion is governed primarily by chain growth, exothermic heat accumulation and light attenuation within the solid volume rather than inhibitor transport from voids. The model correctly identifies that for these massive features, the core material itself is the primary determinant of the final chemical state, rendering the surrounding void less relevant to the prediction.

Diffusion-dominated regimes such as the fine solid regions and struts are represented by dark regions in the heatmap, while the surrounding boundaries or void spaces appear brighter. For high-surface-area structures, the chemical conversion is oxygen-inhibition limited. The degree of conversion in a thin strut is critically dependent on the flux of oxygen diffusing from the immediate surroundings. By masking the void or the interface, we remove the information regarding the oxygen reservoir, which blinds the model to the inhibition potential. Hence, the model has learned that for thin features, the context (the void) is as predictive as the content (the strut). The dark solid indicates that the model looks at the edges to determine the fate of the center, reflecting the physical reality of diffusion-controlled polymerisation.

In every test case, perturbing the phase of importance induces a large shift in prediction, with this shift frequently exceeding 0.010 in magnitude. Conversely, perturbations within the alternate phase elicit a negligible response with a shift of less than 0.002 in magnitude. This order-of-magnitude difference provides evidence that the model is causally linking the solid morphology to the chemical outcome. The model's reasoning adapts correctly to diverse topologies. This robust validation confirms that the C-PGA models aren't just fitting statistical patterns but have learned a generalisable physical principle that the material-void interface is the governing locus of photochemical conversion.

This analysis highlights clearly why occlusion sensitivity is more interpretable here than gradient-based saliency. A gradient map might highlight the solid edges (high frequency), leading to the assumption the model is just doing edge detection. However, occlusion tests reveal a deeper causal logic where the model understands that empty space is physically active. It demonstrates that the network is not merely matching textures but is causally linking the geometry of the void to the chemical conversion of the solid, validating its utility as a physics-aware scientific instrument.

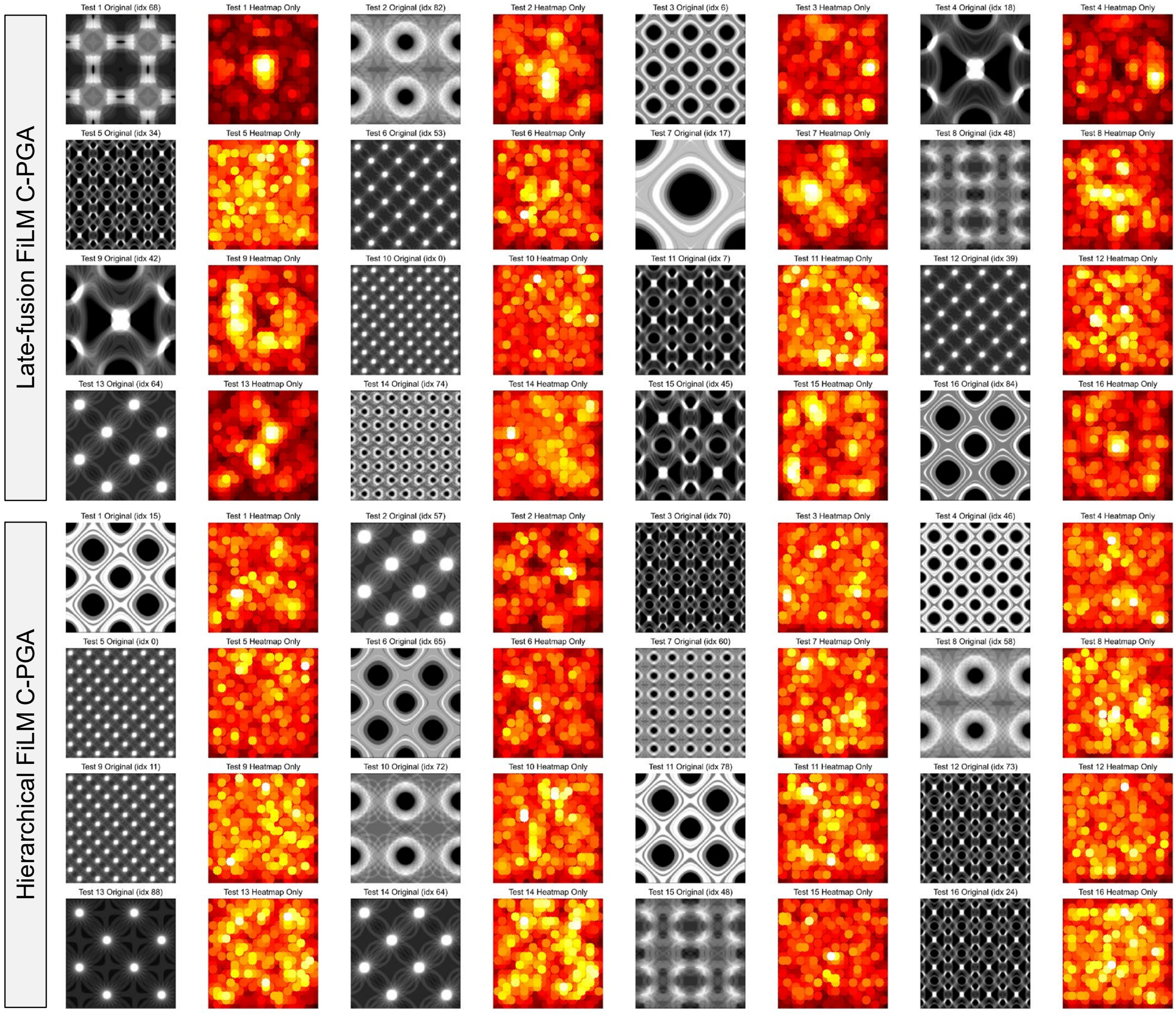

**Figure 5** *Spatial analysis of model uncertainty: Occlusion sensitivity map analysis for 16 random test specimens including for each specimen, the heatmap and the corresponding image slice. Both FiLM-based C-PGA multimodal models are compared side-by-side.*

## 2.6. Interpreting the learned gating parameters for FiLM-based C-PGA models

The primary objective of this work is to move beyond black box predictions and validate that the proposed C-PGA model's reasoning is both trustworthy and physically informed. By analysing the internal parameters of the FiLM layer, we can quantitatively reverse engineer the model's learned logic. Figure 6 plots the relationship between the six input numeric features and the mean $\gamma$ (scaling) parameter that the model learns to apply to the 3D visual feature streams.

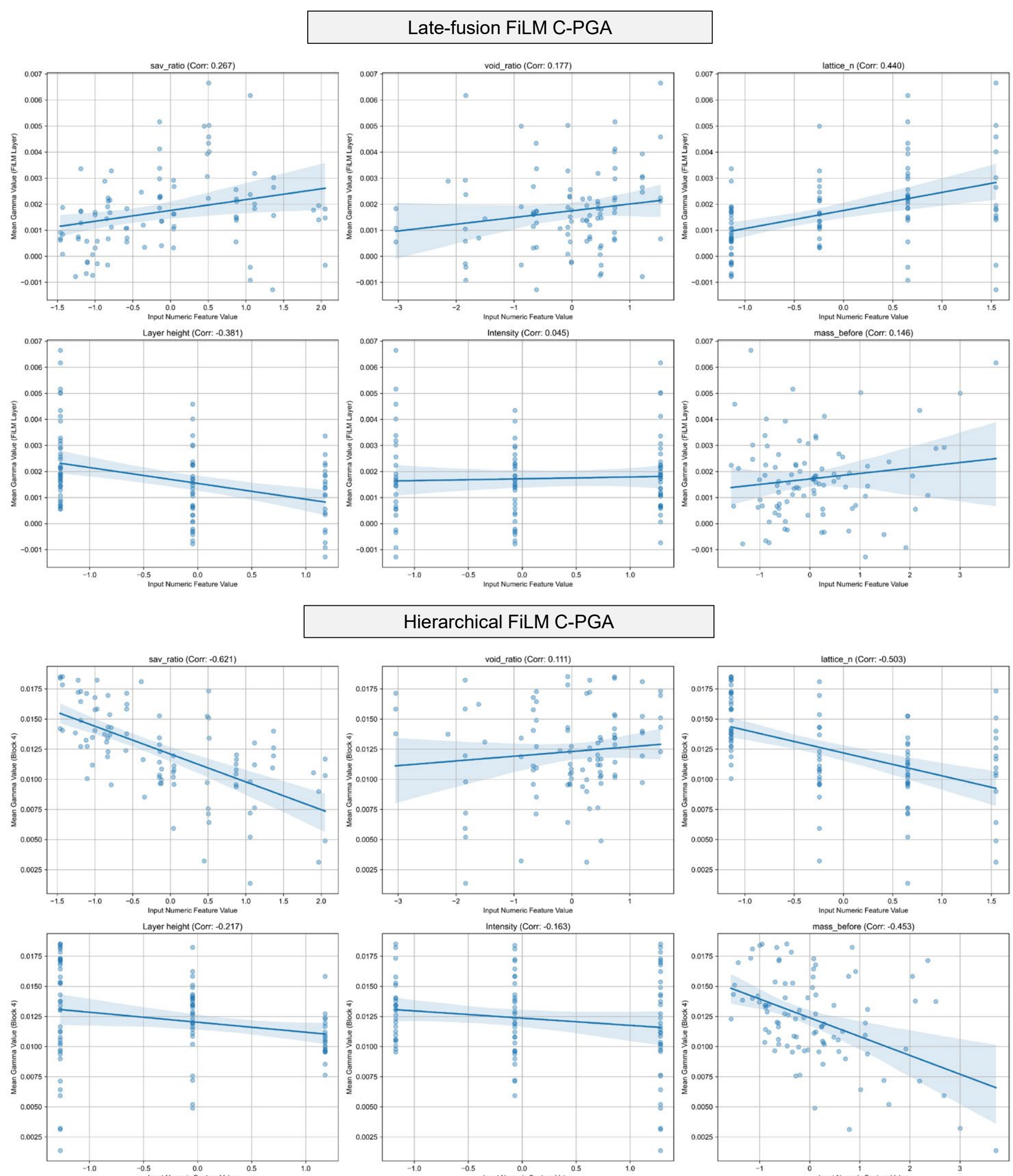

**Figure 6** *Mechanistic interpretation of the learned physics-gating parameters. A quantitative analysis of the internal FiLM layer, plotting the relationship between input physical parameters (normalized) and the learned mean scaling factor ($\gamma$) applied to the visual feature stream. The proposed late-fusion C-PGA model reveals distinct, physically interpretable strategies: it learns to upweight visual features based on geometric topology while systematically dampening visual influence as layer height increases, effectively compensating for increased optical scattering depth. The hierarchical FiLM model exhibits a different logic, dominated by a strong negative correlation with SA/V, indicating that its multi-stage gating prioritises transport-governing parameters. Lack of correlation with intensity across models confirms robustness against overfitting to non-linear variables.*

The results provide evidence of a sophisticated, learned physical strategy rather than spurious correlation. The hierarchical and late-fusion models both learned distinct and physically plausible relationships where the strongest positive correlation for hierarchical was with SA/V ratio (R=-0.621) and for both with lattice_n (R=-0.503 and R=0.440 respectively), confirming that the models' interpretation of visual data is heavily dependent on the fundamental unit cell topology and surface-mediated processes. Even more compelling is the strong negative correlation in the case of late-fusion with layer height (R=-0.391). This demonstrates an intelligent, non-obvious gating as layer height increases (a parameter known to increase light absorption and scattering and reduce the vertical fidelity of the optical projection), the model systematically dampens the influence of the visual feature stream. Conversely, the model correctly identified a negligible direct-scaling relationship for Intensity (R=0.045), proving its robustness against overfitting to non-linear variables that are likely less influential when normalised intensity maps are provided, or are handled by the $\beta$ (shift) parameter.

The analysis performed on the hierarchical FiLM model reveals why it achieved slightly higher predictive fidelity ($R^2$ = 0.7806). Its multi-stage gating process learns a fundamentally different physical logic, dominated by a powerful negative correlation with SA/V (R=-0.621) and mass_before (R=-0.453). This indicates that by modulating features at every spatial scale, the hierarchical model correctly identified that the SA/V a critical parameter governing surface-mediated transport and reaction kinetics is the single most important physical driver. While the hierarchical model captures a deeper representation of the underlying transport physics (as evidenced by its strong SA/V correlation), the late-fusion might be more desirable as detailed in the occlusion analysis. The hierarchical FiLM's distributed reasoning resulted in diffuse and opaque sensitivity maps as the only downside if considering the model for scientific visual interpretability. In contrast, the late-fusion C-PGA provided a comparable high-fidelity prediction ($R^2$=0.7679) while demonstrating clear interpretability.

In this work, we have developed and rigorously validated coupled physics-gated adaptation (C-PGA), a novel deep learning framework for predicting volumetric degree of chemical conversion in additively manufactured lattice structures. By leveraging feature-wise linear modulation (FiLM), the architecture explicitly models the non-linear coupling between sparse physical parameters and dense 3D visual data. This approach effectively integrates idealised geometric data with forward-modelled projection stacks that account for the critical manufacturing physics of optical diffraction and inhibitor diffusion. A defining feature of this study is the prioritisation of scientific trustworthiness over raw predictive metrics. While our ablation study revealed that a complex hierarchical FiLM architecture achieved the highest quantitative scores ($R^2$=0.7806), the late-fusion FiLM model ($R^2$=0.7679, MAE=0.0200) offers superior transparency. Through a rigorous dual-pronged interpretability analysis, we demonstrated that the late-fusion architecture offers a clear window into its decision-making process, whereas the hierarchical approach resulted in opaque, distributed reasoning.

We have transformed the model from a black box predictor into a validated scientific instrument through three key lines of evidence. Spatial uncertainty, where analysis of latency maps reveals that the model's predictive error is not random but correctly co-locates with physically challenging regimes, such as low-porosity lattices and kinetically complex thick-strut geometries. Mechanistic logic, where analysis of the internal gating parameters proves the model has learned valid physical laws such as the inverse relationship between layer height and visual feature reliability, demonstrating an autonomous discovery of optical absorption and scattering limitations. Causal attention, where comparative occlusion sensitivity tests provide ground-truth evidence that the model's spatial reasoning is precisely concentrated at the material-void interface, verifying an attention to regions where diffraction and diffusion effects dominate.

This work establishes a new benchmark for physics-aware deep learning in advanced manufacturing. By demonstrating that FiLM-based physics-gated architectures can achieve state-of-the-art fidelity while remaining interpretable, we provide a blueprint for the next generation of trustworthy digital twins, capable of guiding process optimisation and materials discovery with mechanistic consistency.

# 3. Method

## 3.1. Generating the lattice projection sets

For the training dataset lattice geometries, there is a roughly linear relationship in the same periodic number for surface area and a consistent change for void ratio among all unit cell sizes. As an example, the first selected scaffold was Schwarz Primitive for which the linear regression functions are:

$$\begin{cases} y_{n=1}(P) = 0.3852x - 0.1152, & R^2 = 0.9999 \\ y_{n=2}(P) = 0.1439x - 0.0427, & R^2 = 0.9982 \\ y_{n=3}(P) = 0.0547x - 0.0159, & R^2 = 0.9852 \\ y_{n=4}(P) = 0.0082x - 0.0019, & R^2 = 0.6351 \end{cases}$$

As illustrated in these regression equations, the variable '$x$' represents C value, while '$y$' represents the surface area of primitive lattices. As the value of C increases, the thickness of the material in question also increases, thereby promoting the surface area. It is notable that there is a pronounced decline in the R-squared value when the unit cell reaches four. This can be explained by the equilibrium between pore size and contact area, which has been exceeded, resulting in an uneven tendency. Therefore, it can be concluded that surface area and pore size are influenced by several factors, rather than being solely dependent on the C value. The equation sets for other geometries are provided in Eqns. S18-S21.

All structures were manufactured with a stereolithography printer with an in-plane pixel size of 35 μm. The arylate-based photo-resin was supplied by Multicomp (MP004390). The object bulk DoC was measured through a rigorous solvent extraction method comprising an acetone (>99%) immersion at room temperature under sonication for 1 hour, a consecutive acetone immersion for 23 hours, and solvent evaporation at 70°C for a further 24-hour window. The DoC was calculated using the equation below where $w_1$ represents the mass before solvent extraction and $w_2$ the mass after solvent extraction:

$$Degree\ of\ Conversion\ (DoC) = \frac{w_2}{w_1} \tag{20}$$

The proposed multimodal neural network models designed to predict the DoC, incorporated numerical data, representing manufacturing process and object geometrical parameters, and image data, comprising projection sets and forward-modelled propagation and diffusion maps (Table 1). The architecture employs parallel 3D convolutional neural networks for volumetric object analysis and spatial feature extraction, and a feedforward network for numerical feature encoding. By fusing these modalities, the model leverages complex spatial topology of the printed object alongside governing process parameters to yield a unified model enabling mechanistic understanding of the photochemical material conversion.

<table>
<tr><th colspan="4">Data Modalities</th></tr>
<tr><th colspan="2">Numerical Data</th><th colspan="2">Image Data</th></tr>
<tr><td rowspan="4"><b>Geometrical, and chemical and thermal transport parameters</b></td><td><i>Surface area to volume (SA/V) ratio</i></td><td rowspan="6"><b>Geometry and transport driven print process parameters</b></td><td rowspan="2"><i>Projection slices Transformed</i></td></tr>
<tr><td><i>Porosity (void ratio)</i></td></tr>
<tr><td><i>Lattice unit cell size</i></td><td rowspan="4"><i>Transformed images incorporating inhibitor diffusion and optical diffraction kernel convolution</i></td></tr>
<tr><td><i>Solid volume</i></td></tr>
<tr><td rowspan="2"><b>Print process parameters</b></td><td><i>Layer height</i></td></tr>
<tr><td><i>Max. exposure intensity</i></td></tr>
</table>

**Table 1** *Data modalities used for training the C-PGA model including numerical data, representing manufacturing process and object geometrical parameters, and image data, comprising projection sets and propagation and diffusion maps*

## 3.2. Implementation of diffraction and diffusion

Optical diffraction and molecular diffusion are critical factors that impact the growth of the photo-polymerisation front in all directions. This influence is almost always anisotropic in nature due to physical and chemical boundary conditions imposed on individual voxels by the stereolithographic setup. This in turn manifests itself in a complex nonuniform correlation with geometrical feature length scales. To increase the model's awareness of such physics-induced effects, a point spread function (PSF) is introduced to represent the projection system. The optical PSF was inputted as a 3D Gaussian with different full width at half maximum (FWHM) values for the x and y axes: approximately 0.12 mm and 0.20 mm, respectively. Diffusion of oxygen (which acts as a radical quencher during polymerisation) is also described using a diffusion kernel. The diffusion coefficient (D) for the resin was measured to be 1.5 $\times 10^{-4}$ $mm^2/s$. Hence, the effective diffusion length corresponding to this value, during a typical layer print time of 3 seconds, is approximately 0.05 to 0.1 mm. The combined effects of diffraction and diffusion are modelled using a convolution of the target image with both the PSF and diffusion kernel (Fig 1d). These calculations are converted to a standard deviation (σ) for the Gaussian kernels yielding $\sigma_{x,\ diffraction} = 0.06$, $\sigma_{y,\ diffraction} = 0.08$ and $\sigma_{diffusion} = 0.09$.

## 3.3. The coupled physics-gated model

The model architecture is a physics-aware, spatially modulated network composed of three main components, as shown in Fig 7: (1) Numeric Encoder: The six numerical features are processed by a two-layer feedforward network ($6 \rightarrow 64 \rightarrow 32$), producing a 32-dimensional physics vector ($f_{num}$) This compact vector serves as the control signal for the gating mechanism; (2) Dual 3D convolutional image encoders: two parallel 3D-CNNs are used to process the visual data both original and transformed slices. Each 3D-CNN arm consists of four sequential blocks of 3D convolution, InstanceNorm3d, ReLU, and MaxPool3d, progressively increasing the channel depth ($1 \rightarrow 32 \rightarrow 64 \rightarrow 128 \rightarrow 256$) while down sampling spatial dimensions. Crucially, the outputs of these encoders are not flattened, but are preserved as 3D feature maps ($f_{orig}$ and $f_{conv}$); (3) Feature-wise linear modulation (FiLM) gating: To capture the non-additive coupled interactions between the physical domains, the two 3D feature maps ($f_{orig}$ and $f_{conv}$), are concatenated along the channel dimension to form a single 512-channel visual feature map ($f_{visual}$). The 32-dim physics vector $f_{num}$ is passed through two separate linear projectors to generate a 512-dimensional scaling vector (γ) and a 512-dimensional shifting vector (β). These vectors are broadcast and applied to the visual feature map ($f_{modulated} = \gamma \cdot f_{visual} + \beta$). This FiLM operation allows the numeric physics parameters to spatially modulate the 3D-CNN features, effectively re-weighting visual information on a per-channel basis before a final decision is made. This physics-gated feature map ($f_{modulated}$) is then passed to a final prediction head (1x1 Conv → Flatten → Dropout (p=0.4) → FC (128) → FC (1)) to predict the final degree of conversion.

## 3.4. Alternative models

### 3.4.1. Unimodal baselines

Two unimodal models were trained to establish the lower bound of performance and confirm the necessity of a multimodal approach: (a) Numerical-only: This baseline model was trained using only the six numerical features. The architecture consisted of the numeric encoder (Linear (6, 64) → ReLU → Linear (64, 32)) followed by a simple two-layer prediction head (Linear (32, 16) → ReLU → Linear (16, 1)); (b) Image-only: This model was trained using only the dual 3D image stacks. The numerical branch was removed, and the two 128-dim image feature vectors were concatenated to form a 256-dim vector, which was passed to a prediction head (Linear (256, 128) → ReLU → Linear (128, 1)).

### 3.4.2. Baseline fusion: simple concatenation

This model (concatenated no gate) represents the most common and direct approach to multimodal fusion. It consists of two image encoders: The dual 3D-CNNs process their respective image stacks, ending with a flatten and linear layer to produce two 128-dimensional feature vectors ($f_{orig}$ and $f_{conv}$). A numeric encoder: The MLP processes the six numeric inputs to produce a 32-dimensional feature vector ($f_{num}$). All three vectors are simply concatenated to form a single 288-dimensional vector ($[f_{orig}, f_{conv}, f_{num}]$). This combined vector is passed to a final two-layer head (Linear (288, 128) → ReLU → Linear (128, 1)) for the final regression. This architecture assumes all features are independent and equally important, learning only additive relationships.

### 3.4.3. Attention-based late-fusion architecture

This model tests the hypothesis that visual features should be gated by physical parameters. However, like the concatenation model, it suffers from the '*flatten bottleneck*' as the fusion happens after all spatial information has been destroyed. It includes a multi-head attention mechanism where the two image vectors are concatenated ($f_{orig} + f_{conv}$) to form a 256-dim visual vector ($f_{visual}$). This $f_{visual}$ acts as the Key (K) and Value (V). The 32-dim $f_{num}$ is first projected from 32 to 256 dimensions after which it acts as the Query (Q). This Query attends to the 256-dim $f_{visual}$ (Key/Value) using 8 parallel attention heads computing a 256-dim attention vector ($f_{att}$). A residual connection adds this back: $f_{refined} = f_{visual} + f_{att}$. The final 288-dim vector ($[f_{refined}, f_{num}]$) is passed to the prediction head.

### 3.4.4. Spatially aware feature-wise linear modulation architectures

The feature-wise linear modulation (FiLM) class of models, which includes our proposed architecture, is designed to solve the flatten bottleneck. They apply physics-based gating spatially, before the flatten layer.

Late-fusion FiLM: This model applies gating at a single, late stage. The 3D-CNN encoders run without flattening, producing two 3D feature maps ($f_{orig}$ and $f_{conv}$). These are concatenated into a single 512-channel 3D map ($f_{visual}$). The 32-dim $f_{num}$ vector is passed to two linear layers to generate a 512-dim scaling vector (γ) and a 512-dim shifting vector (β). These are broadcast across the 3D map ($f_{modulated} = \gamma \cdot f_{visual} + \beta$). This single, physics-gated 3D map is passed to the prediction head.

Hierarchical FiLM: This model is our most complex architecture. It injects the numeric physics at every block of the 3D-CNNs. The 3D-CNNs are unrolled into their four constituent blocks. The 32-dim $f_{num}$ vector is fed into 8 different projector networks (a γ/β pair) for each of the four blocks, with channel sizes 32, 64, 128, and 256. After each Conv-Norm-ReLU block, the resulting 3D feature map is immediately modulated by its corresponding γ and β before being passed to the MaxPool3d layer. This allows the numeric physics to guide the visual feature extraction at every spatial scale. The two 256-channel maps are concatenated and passed to the prediction head, just as in late-fusion FiLM.

The full dataset of 648 samples was partitioned into three distinct sets to ensure rigorous, non-overlapping evaluation. First, a 20% hold-out test set (130 samples) was reserved. The remaining 80% (518 samples) was then further split, with 20% of this portion (104 samples) reserved for validation. This resulted in a final data distribution of 64% training, 16% validation, and 20% test. For preprocessing, all numerical features were standardised, with the scaler fit only on the 64% training partition to prevent data leakage. Missing values in the numerical data were imputed prior to scaling.

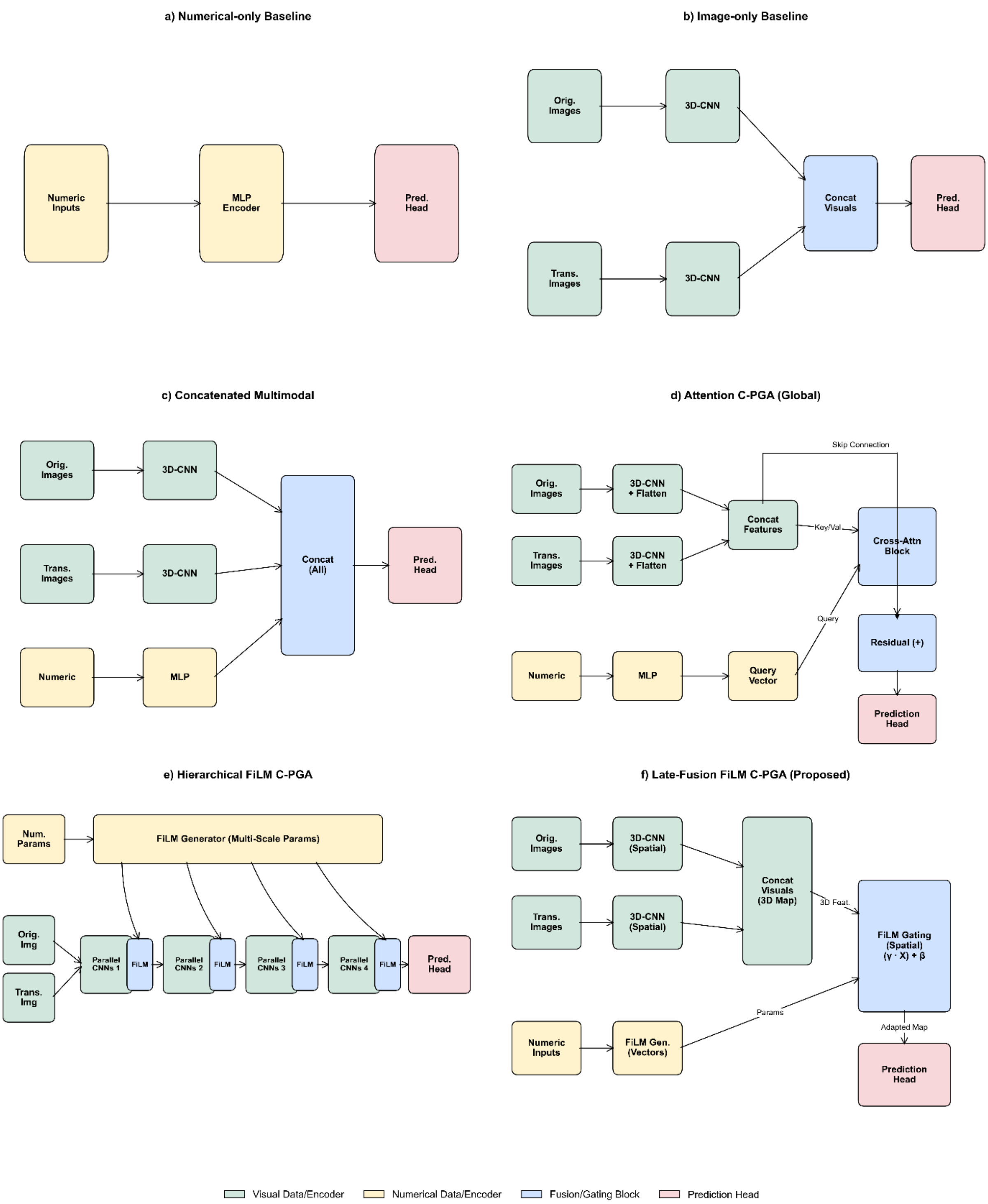

**Figure 7** *Schematic comparison of the six neural network architectures evaluated in the ablation study. The diagrams illustrate the data flow and fusion mechanisms for: (a) Numerical-only baseline; (b) Image-only baseline; (c) Naive concatenation baseline; (d) Attention-based C-PGA (operating on flattened vectors); (e) Hierarchical FiLM C-PGA (multi-scale gating at every block); and (f) Late-fusion FiLM C-PGA. The progression highlights the shift from simple additive fusion (c) to sophisticated physics-informed gating (d-f). The proposed architecture (f) is distinguished by its ability to spatially modulate the 3D feature maps using a single, high-level physics-gating block before flattening, preserving spatial correspondence during the adaptation process. Colour legend: visual data/encoders (green), numerical data/encoders (yellow), fusion/gating blocks (blue), and prediction heads (red).*

The images were grayscale, resized to 160x160 pixels, and normalised. Diffraction was simulated by incorporating an optical point spread function represented as a 3D gaussian with different full width at half maximum (FWHM) values for the x and y axes: averaging at 0.155 mm or a sigma of 0.066. Diffusion of oxygen was modelled by subtracting the convolved inverse of the original image, using a diffusivity of $1.51 \times 10^{-4}$ $mm^2/s$ with a sigma of 0.15, to represent oxygen influx. These convolutions were applied to pre-process images, and the convolved versions were used during training.

A standard mean squared error (MSE) loss is ill-suited for this problem, as the physical cost of an error is highly non-linear. For example, a predicted conversion of 0.7 versus an actual 0.6 represents a far more critical engineering failure at the extremes of what may be acceptable, than a prediction of 0.9 versus 0.8. Hence, while there are fewer samples here, these samples are extremely important to the model. To encode this domain knowledge directly into the model, a graduated weighted MSE (wMSE) loss function was implemented. This function applies a standard weight (1.0) for acceptable-to-high conversion samples (target > 0.87) but aggressively increases the penalty on errors in lower-conversion regions, which represent critical failures (7.0 for target below 0.6, 6.0 for target in [0.6, 0.7), 5.0 for target in [0.7, 0.8), and 4.0 for target in [0.8, 0.87)). This cost-sensitive approach forces the model to strongly prioritise the avoidance of under-conversion. For training, a step scheduler was used to reduce the learning rate by a factor of 0.9 every 10 epochs, promoting stable convergence.

This multimodal approach integrates engineered numerical features with spatial image features using two dedicated 3D CNNs, followed by concatenation and a unified prediction layer. By combining convolutional features from both original and convolved images, the model captures detailed spatial representations, while the numerical features provide critical material-specific insights. The inclusion of diffraction and diffusion modeling via Gaussian convolution helps simulate real-world conditions, improving the robustness of the model's predictions on the DoC.

## 4. Author contributions

H.H. conceived the project and designed the research. M.E. and H.H. developed the C-PGA deep learning architectures, implemented the FiLM gating mechanisms, and performed the computational analyses. H.H., C.Z., J.W. and X.C. designed geometries and performed the 3D printing experiments to produce the additively manufactured specimen dataset and chemical characterization. All authors were involved in writing the main manuscript text.

## 5. Competing interests

The authors declare no competing interests.

## 6. Acknowledgements

The authors would like to thank the UK Research and Innovation (UKRI). This work was supported by the UKRI Future Leaders Fellowship [grant number MR/X034976/1]

## 7. Code and data availability

The code used to process the raw data, implement the C-PGA architecture and train the models is available at [https://anonymous.4open.science/r/C-PGA-6A78/](https://anonymous.4open.science/r/C-PGA-6A78/). The dataset experimentally acquired and analysed during the current study is available from the corresponding author on reasonable request.

## Coupled Physics-Gated Adaptation: Spatially Decoding Volumetric Photochemical Conversion in Complex 3D-Printed Objects

## Supplementary Information

### 1. Equations for generation of TPMS structures

The most convenient and prevalent method for defining various triply periodic minimal surfaces is through the construction of specific level-set approximation equations. These are a series of trigonometric functions that align with the catenoid as a ground law $f(x, y, z) = c$ which can be expressed as follows: In this function, the variables x, y, and z represent the collection of points in the given three-dimensional area. The function $f(x, y, z)$ describes the minimal surface formed by points that satisfy a specific formula as we explained before [9]. Additionally, introducing $c$ as the isosurface value allows for the defined controlled distance between the generated surface and the actual theoretical minimal surface. When the isovalue equals zero, the minimal surface forms a topology firm. The presented work primarily targets the following lattices and their respective level-set approximation equations [1-6].

Schwarz Primitive:

$$f(x, y, z) = cos(x) + cos(y) + cos(z) + c \qquad (S1)$$

Schwarz Diamond:

$$f(x, y, z) = cos(x)\,cos(y)\,cos(z) - sin(x)\,sin(y)\,sin(z) + c \qquad (S2)$$

Fischer-Koch:

$$f(x, y, z) = cos(2x)sin(y)cos(z) + cos(2y)sin(z)cos(x) + cos(2z)sin(x)cos(y) + c \qquad (S3)$$

Gyroid:

$$f(x, y, z) = sin(x)\,cos(y) + sin(z)\,cos(x) + sin(y)\,cos(z) + c \qquad (S4)$$

Neovius:

$$f(x, y, z) = 3\big(cos(2x) + cos(2y) + cos(2z)\big) + 4cos(2x)cos(2y)cos(2z) + c \qquad (S5)$$

F-RD:

$$f(x, y, z) = 4cos(x)cos(y)cos(z) - [cos(2x)cos(2y) + cos(2x)cos(2z) + cos(2y)cos(2z)] + c \qquad (S6)$$

The marching cube algorithm is employed in conjunction with the aforementioned functions to construct constant density isosurfaces. All points from the given set of coordinates $(x, y, z)$ are used to determine a plane through three points, forming a multitude of triangles. Subsequently, the data points are processed by the algorithm in three-dimensional scan-line order, and then triangle vertices are stored using linear interpolation to create a smooth surface. The generation of a higher-resolution system results in the formation of a surface with a more uniform and realistic appearance. Each of these equations represents one distinctive TPMS architecture, *x, y, z* normally incorporated with topology parameters during the practical design phase. However, due to the inherent nature of trigonometric functions, these equations can be converted or transformed into alternative formats while maintaining the overall structural integrity.

## 2. Building structures from surfaces

As previously stated, the c value, which can be described as the offset parameter, defines the distance between the central minimal surface and the generated one. This will result in the given boundary spatial area being divided into two separate areas. Points that occupy areas above the surface, outside the structure, are represented by $f(x,y,z) > c$, whereas points that occupy areas below the surface, inside the structure, are represented by $f(x,y,z) < c$. In both cases, the value of c represents the distance between the central minimal surface and the generated one. The outlined level-set mathematical approximation formulas define TPMS lattices as minimal surfaces, thereby limiting their applicability to the film level. It is therefore necessary to construct TPMS scaffolds with a thickness. In the existing literature, two general approaches to creating a self-supporting configuration are identified: the skeletal and sheet structure. These are otherwise known as the solid and network TPMS structures. The skeletal methods typically generate thickness by filling the occupied area between the isosurface and the set boundary, where $f(x,y,z) < c$. Based on the built method, skeletal-based TPMS structures also hold features that are defined by the boundary geometry, the most common situation is constructed inside the classic crystalline form. When the offset parameter is set to zero, the two spaces are of equal volume and have a relative density of one. Consequently, the offset parameter initially affects the relative density through the system and subsequently begins to influence the thickness. To obtain the self-supported skeletal materials, some of the complicated graphs are defined by this ground equation and some as shown below:

$$F(x,y,z) = a_1T_1 + a_2T_2 + t \tag{S7}$$

$$F(Gyroid) = 10(cosxsinx + cosysinz + coszsinx) - 0.5(cos2xcos2y + cos2ycos2z + cos2zcos2x) + t \tag{S8}$$

$$\begin{aligned} F(Diamond) = {} & 10\big(sin(x-4\pi)sin(y-4\pi)sin(z-4\pi) + sin(x-4\pi)cos(y-4\pi)cos(z-4\pi) \\ & + cos(x-4\pi)sin(y-4\pi)cos(z-4\pi) + cos(x-4\pi)cos(y-4\pi)sin(z-4\pi)\big) \\ & - 0.7(cos4x + cos4y + cos4z) + t \end{aligned} \tag{S9}$$

$$F(Primitive) = 10(cosx + cosy + cosz) - 5.1(cosxcosy + cosycosz + coszcosx) + t \tag{S10}$$

Another widely adopted method for the construction of sheet-based TPMS scaffolds is the offset of the minimal surface in two opposing directions. This results in the generation of parallel surfaces that can subsequently form a solid structure with defined boundary lines. In general, this approach is processed by a set of one positive and one negative, with the same absolute c value used to restrict them inside sheets. This is expressed as $-c < f(x,y,z) < c$. In contrast to the preceding method, the sheet-based approach enables direct alteration of the thickness by modifying the c value. The theoretically generated thickness is expected to be approximately twice the isovalue, beyond the c value exceeding the specified range, the entire system will reach to collapse. However, the thickness is influenced by both the unit cell geometry and the offset parameter, and thus further investigation or characterisation of the generated thickness is required.

The normal vector approach requires further calculation of the direction vector and offset distance shown in the equations. Different from two opposite constant c values, here we use a function including three-dimensional factors to optimise a fixed thickness to construct the structures. It has been also evidently indicated that this innovative method can enhance both the mechanical performance and the fatigue life of the material.

$$f_1(x,y,z) = f(x,y,z) + c(x,y,z) \quad \text{(S11)}$$

$$f_2(x,y,z) = c(x,y,z) - f(x,y,z) \quad \text{(S12)}$$

$$c(x,y,z) = c_0(x,y,z) \cdot (\, t(x,y,z) \,/\, t_0(x,y,z), 6 \,) \quad \text{(S13)}$$

The operation below was used to construct the scaffolds.

$$F_1 = F + C \quad \text{(S14)}$$

$$F_2 = F - C \quad \text{(S15)}$$

To impose constraints on the structures located within these offset surfaces,

$$F = min(F_1, F_2) \quad \text{(S16)}$$

Emerging this topology parameter with their origin equations, modified specific functions. Where $X=2\pi Tx$, $Y=2\pi Ty$, $Z=2\pi Tz$, T represents the repeated periodic numbers, and $(x, y, z)$ was assigned a resolution of 100 in all three dimensions. One of the generation processes as examples can be shown below.

Schwarz Primitive:

$$f(x,y,z) = cos(2 \times \pi \times T \times x) + cos(2 \times \pi \times T \times y) + cos(2 \times \pi \times T \times z); \quad \text{(S17)}$$

$$F_1 = F + c;$$

$$F_2 = c - F;$$

$$F(P) = min(F_1, F_2);$$

The regulated constant c value and repeating times T thus define the relative density, porosity, and surface area in a simultaneous manner.

**Table S1** *The configuration parameters of each print thickness*

| Layer thickness (mm) | Exposure time (s) | Initial exposure time (s) | Layer numbers (total) | Platform velocity (mm/s) | Turn off delay (s) | Rising height (mm) | Base exposure layers |
|---|---|---|---|---|---|---|---|
| 0.05 | 2.6 | 40 | 200 | 1 | 3 | 5 | 2 |
| 0.10 | 3 | 40 | 100 | 1 | 3 | 5 | 2 |
| 0.15 | 4 | 50 | 67 | 1 | 3 | 5 | 2 |

The initial exposure time represents the exposure time of each base layer, which is set to two layers for the purposes of providing enhanced support. Subsequently, the exposure time, defined as the duration of each layer being illuminated, is a configuration-dependent factor that typically involves a trial-and-error process and varies across scenarios. Subsequently, the scan speed, turn-off delay and rising height are maintained at consistent levels across varying thicknesses, as these variables have been classified as irrelevant in this context.

## 3. Measuring surface area to volume ratio and porosity

The surface area of the cellular lattice structures was estimated using the triangle vertices calculation, which employed the same mechanism as the isosurface mesh method. This involved the generation of point sets and the subsequent theoretical vertices. The methodology initially entails a comprehensive scan search of all points, after which each point forms a mesh triangle. The surface area of each triangle is then calculated based on the coordinates of the points, with the total area determined by accumulated surface area. This method is frequently employed in vector geometry for the purpose of determining the area of a triangle within a three-dimensional space. Alternatively, the solid volume was calculated utilising the voxel stacking method, which initially divided the three-dimensional space into a multitude of voxels, each defined by a specific solution. In a similar manner, the generated points are scanned and searched; if the calculated value is positive, 1 is returned for the defined voxel of the point, whereas if the calculated value is negative, zero is returned. The final step is to stack all the individual dimensional voxel products to obtain the solid volume.

Moreover, all calculated surface areas and solid volumes have been verified by the 3D processing software Blender and Halot Box. It is noteworthy that the two different software-implemented algorithms for geometry parameters are different. The former employed a vector-based methodology, whereas the latter opted to utilise a voxel-based approach, thereby pre-stimulating the consumption of resins and enhancing the accuracy of the practical situation. The results demonstrated that the margin of error was approximately 1.5%, indicating a successful calculation.

## 4. Equations correlating surface area to C value

Diamond (Eqn. S18):

$$\begin{cases} y_1(D) = 2.1176x + 7.998, \quad R^2 = 1.9955 \\ y_2(D) = 0.0440x + 16.1118, \quad R^2 = 0.1700 \\ y_3(D) = -2.0513x + 24.2688, \quad R^2 = 0.9489 \\ y_4(D) = -4.1925x + 32.4941, \quad R^2 = 0.9762 \end{cases}$$

In the case of diamond lattices, the impact of the c value is more apparent. When the geometry passes the balance point, as previously demonstrated, the surface area tends to have a linear negative correlation with the offset parameter, resulting in a decrease as the boundary thickens. This discovery corroborates the existing literature, which suggests that the large surface area of diamond lattices, may indicate that this category of TPMS structures has more potential to be utilised in energy absorption applications following further precise optimisation.

Gyroid (Eqn. S19):

$$\begin{cases} y_1(G) = 2.4453x + 6.3182, & R^2 = 1.9995 \\ y_2(G) = 1.4886x + 0.0427, & R^2 = 1.9902 \\ y_3(G) = 0.5258x + 19.0995, & R^2 = 0.8750 \\ y_4(G) = -0.4495x + 25.5265, & R^2 = 0.7241 \end{cases}$$

The results indicate that the linear function applied may not be an optimal means of explaining the Gyroid structure. This may reflect the complex curvilinear orientation inherent to Gyroid structures. Moreover, the critical equilibrium point of the gyroid structure was observed to lie between unit numbers 3 and 4.

Neovius (Eqn. S20):

$$\begin{cases} y_1(N) = 0.4815x + 7.8968, & R^2 = 1.9999 \\ y_2(N) = 0.1013x + 14.9676, & R^2 = 1.9994 \\ y_3(N) = -0.2642x + 22.0328., & R^2 = 0.9938 \\ y_4(N) = -5.34x + 31.7281, & R^2 = 0.9963 \end{cases}$$

The final selective TPMS is Neovius, in which the thickness also manifested as a linear regression following the boundary point. However, in comparison to the preceding TPMS scaffolds, the thickness variation across different c values at identical periodic conditions appears to be more nuanced, bearing resemblance to the primitive structures. This discrepancy may be attributed to the shared organised porous pattern between the Primitive and Neovius geometries, which differs from that observed in other structures. Following the analysis of the void ratio, the porosity was altered by modifying the C value. However, the unit cell size of each TPMS lattice with the same c value remained constant. As a result, the focus shifted to a comparison of different geometries (Eqn. S21).

$$\begin{cases} y_{\text{Primitive}} = -0.567x + 0.9998, \; R^2 = 0.99998 \\ y_{\text{Diamond}} = -0.8255x + 1.0037, R^2 = 0.99997 \\ y_{\text{Gyroid}} = -0.6475x + 1.0017, \; R^2 = 0.99998 \\ y_{\text{Neovius}} = -0.1815x + 0.8115, \; R^2 = 0.99957 \end{cases}$$

In this context, the variable x represents the C value, while y denotes the void ratio, which has been previously stimulated. It is evident that all selected geometries exhibit a linear relationship between the offset parameter and porosity. Furthermore, the absolute value of the regression coefficient in Diamond is the largest, indicating that the geometry properties of Diamond will be influenced by c to the greatest extent. The largest surface area of Diamond, which can reach 30 $cm^2$, is the highest among all selected structures. In contrast, the Neovius remains relatively inconsequential or limited by the actual setting, consistent with what has also been observed previously.

## 5. Microstructural morphology

Following the acetone treatment for a period of 24 hours, most of the samples exhibited the presence of notable cracks or fractures on the surface, with some instances displaying a penetration through the entirety of the structure as the network weakens and defects arise. Consequently, a selection of standard samples, exhibiting identical geometries but varying printing layer thicknesses, was subjected to scanning electron microscopy (SEM) (TM4000Plus, HITACHI) to facilitate the observation of the microstructure of the fracture parts. Since morphological characterisation can be conducted on a remotely large field in comparison to the conventional elaborate film, the TPMS samples did not undergo sputtering, eliminating the necessity for enhanced second electron emission to facilitate a superior view. Instead, they were placed in a vacuum environment in an ordered sequence, enabling direct observation of the microstructure across diverse structures. In accordance with the agreed account of the aforementioned information, the images were obtained in three progressively increased magnifications 30x, 60x, and 300x for all six TPMS lattices in two geometry configurations, conducted by three different layer heights, as Figure S1 demonstrates below.

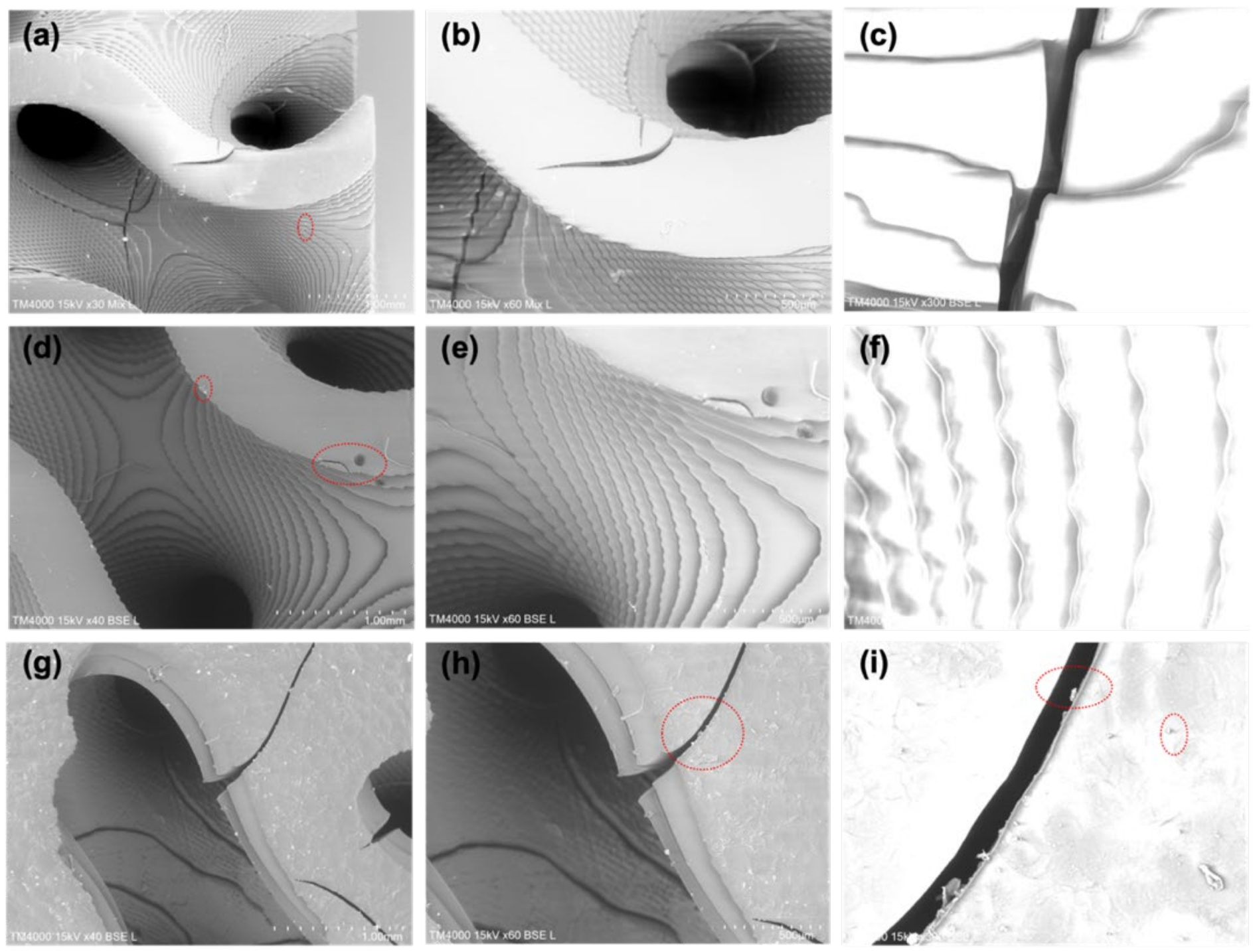

**Figure S1** *The SEM images for Gyroid (unit cell = 2) TPMS scaffolds. (a) 50 µm layer height, 30x magnification. (b) 50 µm layer height, 60x magnification. (c) 50 µm layer height, 300x magnification. (d) 100 µm layer height, 40x magnification. (e) 100 µm layer height, 60x magnification. (f) 100 µm layer height, 300x magnification. (g) 150 µm layer height, 40x magnification. (h) 150 µm layer height, 60x magnification. (i) 150 µm layer height, 300x magnification.*

As illustrated in the preceding section, the SEM images of the identical structure, the gyroid with a periodic repetition of two times, but sliced with disparate layer heights (following model configuration in Figure S2) and observed with varying magnifications, demonstrate that as the layer thickness is increased, the alteration of the stack layer becomes more evident, while also exhibiting a notable increase in the fractured part among the structure. The small bright points visible in the images indicate the presence of residual monomers that were not fully extracted during the post-processing stage. These monomers may have existed internally or remained on the surface. It is also evident that an increase in layer thickness results in a greater prevalence of interlayer pores, which may ultimately lead to material failure and unintended fracture. However, these fractures would not contribute to surface area changes during solvent extraction as they only appear due to mechanical stresses during the drying process.

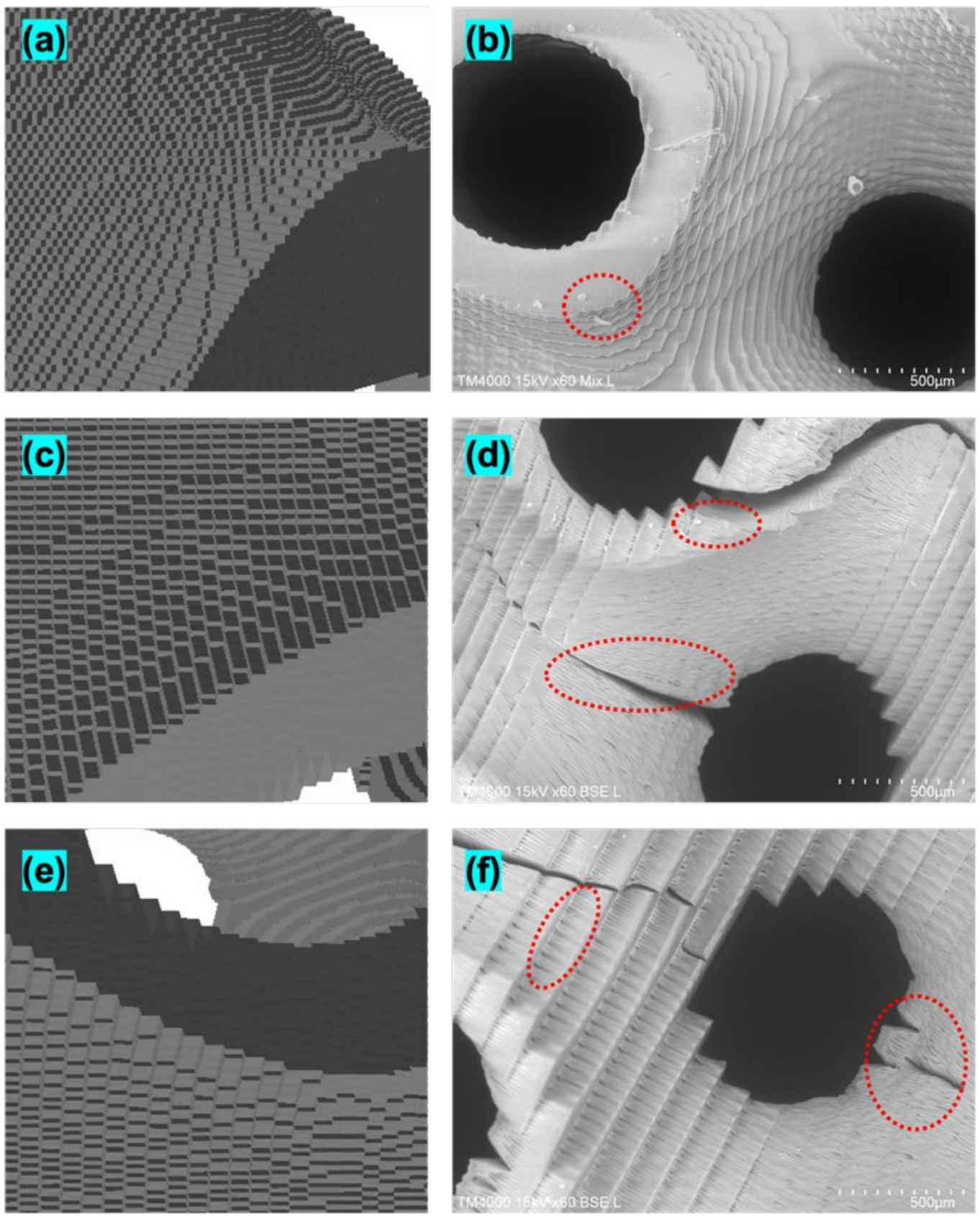

**Figure S2** *The comparison of sliced single Primitive meshed close structure and the SEM images of Primitive (unit cell=4) of 60x magnification. (a-b) 50 µm. (c-d)100 µm. (e-f) 150 µm.*

A high level of coherence can be readily discerned when the close shot of layered print samples is compared with the structures observed in the SEM field (model vs real print in Figure S2). Although the single primitive structures were implemented for this detailed comparison with the periodic repeating pattern at four, where the pore size is at different levels, they still provide strong guidance for our analysis. The boundary edge of each layer is worthy of further investigation and analysis. As the layer thickness increases, the surface becomes more rough and uneven. The orientation of distinct rectangular solid structures following the pixels of the projected image, is also evident in these SEM images